\documentclass[11pt,a4paper]{article}
\usepackage[hyperref]{emnlp2020}
\usepackage{times}
\usepackage{latexsym}

\usepackage{microtype}

\aclfinalcopy 

\usepackage{arydshln}
\usepackage{times}
\usepackage{latexsym}
\usepackage{url}
\usepackage{helvet}
\usepackage{courier}
\usepackage{graphicx}
\usepackage{adjustbox}
\usepackage{amsmath,amssymb,amsthm}
\usepackage{mathabx}
\usepackage{multirow}
\usepackage{hhline}
\usepackage{subcaption}
\usepackage{enumerate}

\usepackage{pgfplots}
\pgfplotsset{compat=1.13}

\definecolor{findOptimalPartition}{HTML}{D7191C}
\definecolor{storeClusterComponent}{HTML}{FDAE61}
\definecolor{dbscan}{HTML}{ABDDA4}
\definecolor{constructCluster}{HTML}{2B83BA}

\usepackage{pgfplots}
\usepackage{xcolor}
\usetikzlibrary{arrows,intersections}

\usepackage{filecontents}
\usepgfplotslibrary{dateplot}

\newcommand*\circled[1]{\tikz[baseline=(char.base)]{
            \node[shape=circle,draw,inner sep=0.001pt] (char) {#1};}}

\newcommand*{\affaddr}[1]{#1} 
\newcommand*{\affmark}[1][*]{\textsuperscript{#1}}

\newcommand\mdoubleplus{\ensuremath{\mathbin{+\mkern-10mu+}}}

\usepackage{blindtext}
\newcommand\blfootnote[1]{%
	\begingroup
	\renewcommand\thefootnote{}\footnote{#1}%
	\addtocounter{footnote}{-1}%
	\endgroup
}

\title{
TopicBERT for Energy Efficient Document Classification
}

\author{*Yatin Chaudhary\affmark[1,2], *Pankaj Gupta\affmark[1],\\
  {\bf Khushbu Saxena\affmark[3], Vivek Kulkarni\affmark[4],
  Thomas Runkler\affmark[3], 
  Hinrich Sch\"{u}tze\affmark[2]}\\ 
 \affaddr{
 {\affmark[1]DRIMCo GmbH, Munich, Germany} $|$ 
 {\affmark[2]CIS, University of Munich (LMU), Munich, Germany} \\
 {\affmark[3]Siemens AG, Munich, Germany} $|$
 \affmark[4]Stanford University, California, USA }\\
  {\tt {yatin.chaudhary}@drimco.net}
}

\date{}

\begin{document}
\maketitle

\begin{abstract}
Prior research notes that BERT's computational cost grows quadratically with sequence length thus leading to longer training times, higher GPU memory constraints and carbon emissions. While recent work seeks to address these scalability issues at pre-training, these issues are also prominent in fine-tuning especially for long sequence tasks like document classification. Our work thus focuses on optimizing the computational cost of fine-tuning for document classification. We achieve this by \emph{complementary learning} of both topic and language models in a unified framework, named {\it TopicBERT}. 
This significantly reduces the number of self-attention operations -- a main performance bottleneck. Consequently, our model achieves a 1.4x ($\sim40\%$) speedup 
with  $\sim40\%$ reduction in CO\textsubscript{2}  emission 
while retaining 99.9\% performance over 5 datasets. 

\end{abstract}

\section{Introduction}
\blfootnote{*Equal Contribution}
Natural Language Processing (NLP) has recently witnessed a series of breakthroughs by the evolution of large-scale language models (LM) such as ELMo \cite{ELMO2018}, BERT \cite{BERT2019}, RoBERTa \cite{RoBERTa2019}, XLNet \cite{xlnet2019} etc. due to improved capabilities for language understanding \cite{BengioDVJ03, MikolovSCCD13}. However this massive increase in model size comes at the expense of very high computational costs: 
longer training time, high GPU/TPU memory constraints, adversely high carbon footprints, 
and 
unaffordable invoices for small-scale enterprises. 

\begin{figure}[t]
  \centering
  \includegraphics[scale=0.34]{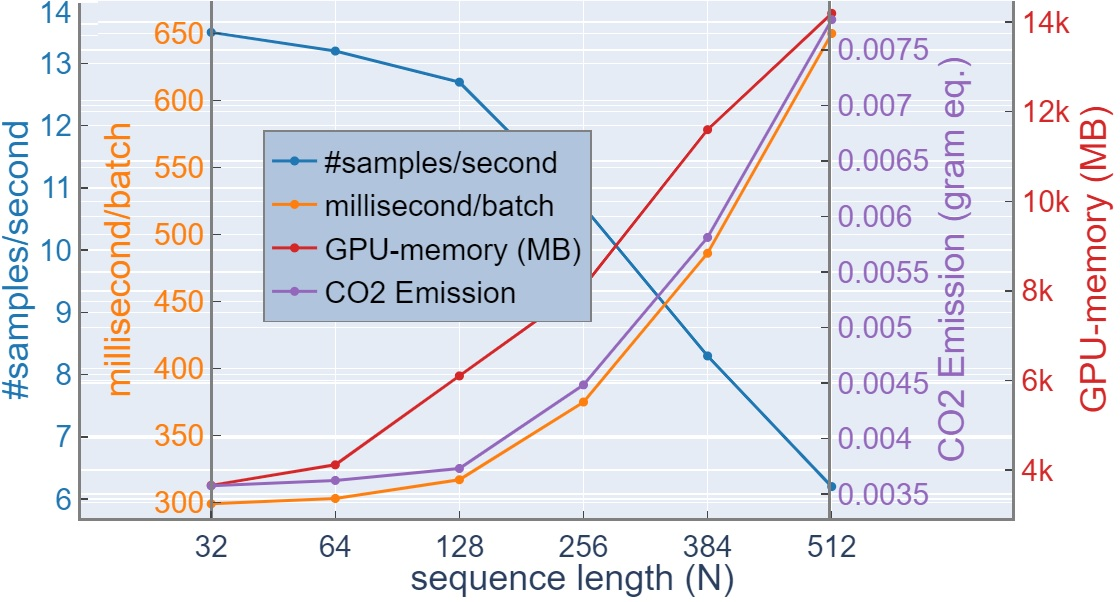}
  \caption{Computational cost vs sequence length}
  \label{fig:motivation}
\end{figure}

Figure \ref{fig:motivation} shows the computational cost (training time: millisecond/batch; CO\textsubscript{2} emission, and GPU memory usage) of BERT all of which grow quadratically with sequence length (N). We note that this is primarily due to self-attention operations. Moreover, as we note in Table 1, the staggering energy cost is not limited to only the {\it pre-training} stage but is also encountered in the fine-tuning stage when processing long sequences as is needed in the task of document classification. Note that the computational cost incurred can be quite significant especially because fine-tuning is more frequent than pre-training and BERT is increasingly used for processing long sequences. Therefore, this work focuses on reducing computational cost in the {\it fine-tuning} stage of BERT especially for the task of document classification.

\begin{table}[t]
		\centering
		\resizebox{0.46\textwidth}{!}{
			\begin{tabular}{lr}
				\hline
				& \multicolumn{1}{c}{\bf CO\textsubscript{2}} \\ 
				BERT pre-training (NAS) \cite{DBLP:conf/acl/StrubellGM19}	&		626k			\\
				BERT fine-training (n=512)*								&		+ 125k \\ \hline
		\end{tabular}
		}	
		\caption{Similar to \newcite{DBLP:conf/acl/StrubellGM19} who estimate the carbon footprint of BERT during pretraining, we estimate the carbon footprint (lbs of CO\textsubscript{2} equivalent) during finetuning BERT for document classification.  *:  see \textit{supplementary} material for details.
}
		\label{tab:co2footprints}
\end{table}

Recent studies address the excessive computational cost of large language models (LMs) in the pre-training stage using two main compression techniques:
(a) {\it Pruning} \cite{MichelLN19, ALBERT:ICLR2020} by reducing model complexity, and 
(b) {\it Knowledge Distillation} 
\cite{HintonKD2015, Tang2019, Turc2019, DistilBERT2019} 
which a student model (compact model) is trained to reproduce a teacher (large model) leveraging the teacher's knowledge.   
Finally, in order to process long sequences, \newcite{xie2019unsupervised} and \newcite{JoshiLZW19}
investigate simple approaches of truncating or partitioning them into smaller sequences,  
e.g., to fit within 512 token limit of BERT for classification; 
However, such partitioning leads to a loss of discriminative cross-partition information 
and is still computationally inefficient. In our work, we address this limitation by learning a complementary 
representation of text using topic models (TM)) \cite{DBLP:journals/jmlr/BleiNJ03, DBLP:conf/icml/MiaoYB16, DBLP:conf/aaai/GuptaCBS19}. 
Because topic models are bag-of-words based models, they are more computationally efficient than large scale language models that are contextual. 
Our work thus leverages this computational efficiency of TMs for efficient and scalable fine-tuning for BERT in document classification.

Specifically our contributions
{\bf (1) Complementary Fine-tuning}:
We present a novel framework: {\it TopicBERT}, 
i.e., topic-aware BERT that leverages the advantages of both neural network-based TM 
and Transformer-based BERT to achieve an improved document-level understanding. 
We report gains in document classification task with full self-attention mechanism and topical information. 
{\bf (2) Efficient Fine-tuning}: {\it TopicBERT} offers an efficient fine-tuning of BERT for long sequences by reducing the number of 
self-attention operations and jointly learning with TM.  We achieve a 1.4x ($\sim$ $40$\%) speedup while retaining 99.9\% of 
classification performance over 5 datasets.
Our  approaches are {\it model agnostic}, therefore we extend BERT and DistilBERT models. Code in available at \url{https://github.com/YatinChaudhary/TopicBERT}.

{\bf Carbon footprint ($CO_2$) estimation}: We follow~\citet{DBLP:journals/corr/abs-1910-09700} and use ML CO\textsubscript{2} Impact calculator\footnote{\url{https://mlco2.github.io/impact/}} to estimate the carbon footprint ($CO_2$) of our experiments using the following equation:

\begin{equation*}
    \begin{aligned}
        CO\textsubscript{2} =&~\mbox{Power consumption} \times \mbox{Time (in hours)} \\
        &\times \mbox{Carbon produced by local power grid}
    \end{aligned}
\end{equation*}


where,
Power consumption = 0.07KW for NVIDIA Tesla T4 16 GB Processor and 
Carbon produced by local power grid = 0.61 kg CO\textsubscript{2}/kWh. 
Therefore, the final equation becomes:

\begin{equation}
    \begin{aligned}
        CO\textsubscript{2} =&~0.07 kW \times \mbox{Time (in hours)} \\
        &\times 0.61 \times 1000 \mbox{ gram eq. CO\textsubscript{2}/kWh}
    \end{aligned}
    \label{eq:co2_estimation}
\end{equation}

In Figure 1, we run BERT for different sequence lengths (32, 64, 128, 256 and 512) with batch-size=4 to estimate GPU-memory consumed and $CO_2$ using equation~\ref{eq:co2_estimation}. We run each model for 15 epochs and compute run-time (in hours).

For Table 1, we estimate $CO_2$ for document classification tasks (BERT fine-tuning) considering 512 sequence length. We first estimate the total BERT fine-tuning time in terms of research activities and/or its applications beyond using multiple factors. Then, using equation~\ref{eq:co2_estimation} the final $CO_2$ is computed. (See \textit{supplementary} for detailed computation)



\begin{figure}[t]
  \centering
  \includegraphics[scale=0.96]{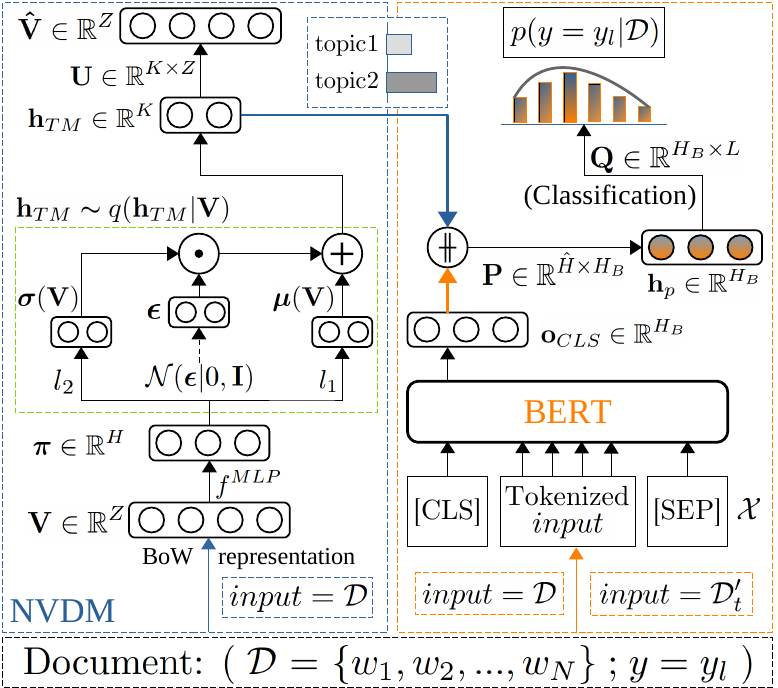}
	\caption{Topic-aware BERT ({\it TopicBERT}):  Joint fine-tuning of NVDM and BERT; 
The {\it input} in BERT is ${\mathcal D}$ for complementary fine-tuning while 
${\mathcal D}'_t$ ($t^{th}$ partition of ${\mathcal D}$) for complementary+efficient fine-tuning.
$\oplus$: addition; $\odot$: Hadamard product;
{\small \circled{\mdoubleplus{}}}: concatenation;
Green dashed lines: variational component of NVDM.} 
	\label{fig:topicBERTarchitecture}
\end{figure}

\section{Methodology: TopicBERT}

Figure~\ref{fig:topicBERTarchitecture} illustrates the architecture of {\it TopicBERT} consisting of: 
(1) Neural Topic Model (NTM),  
(2) Neural Language Model (NLM) 
to achieve complementary and efficient document understanding. 

\subsection{TopicBERT: Complementary Fine-tuning}

Given a document $\mathcal{D}$ = $[w_1,..., w_N]$ of sequence length $N$, consider $\mathbf{V}$ $\in$ $\mathbb{R}^{Z}$ be its BoW representation,  
$\mathbf{v_i}$ $\in$ $\mathbb{R}^{Z}$ be the
one-hot representation of the word at position $i$ and $Z$ be the vocabulary size.

The \textbf{Neural Topic Model} component (Figure~\ref{fig:topicBERTarchitecture}, left)
is based on Neural Variational Document Model (NVDM) \citep{DBLP:conf/icml/MiaoYB16}, seen as 
a variational autoencoder for document modeling in an unsupervised generative fashion such that:

(a) an MLP {\it encoder} $f^{MLP}$ and two linear projections $l_1$ and $l_2$ 
compress the input document $\mathbf{V}$ into a continuous hidden vector $\mathbf{h}_{TM} \in \mathbb{R}^K$:\vspace{-0.3cm}

{\small  \begin{equation*}\label{eq:nvdm}
	\begin{aligned}
		&\boldsymbol{\pi} = g(f^{MLP}(\mathbf{V}))\;\; \mbox{and}\;\;\boldsymbol{\epsilon} \thicksim \mathcal{N}(0,\mathbf{I}) \\
		&\boldsymbol{\mu}(\mathbf{V}) = l_1(\boldsymbol{\pi})\;\;\;\;\;\;\;\;\mbox{and}\;\;\boldsymbol{\sigma}(\mathbf{V}) = l_2(\boldsymbol{\pi}) \\
		&q(\mathbf{h}_{TM}|\mathbf{V}) = \mathcal{N}(\mathbf{h}_{TM}|\boldsymbol{\mu}(\mathbf{V}), \mbox{diag}(\boldsymbol{\sigma}(\mathbf{V}))) \\
		& \mathbf{h}_{TM} \thicksim q(\mathbf{h}_{TM}|\mathbf{V})   \implies \mathbf{h}_{TM} = \boldsymbol{\mu}(\mathbf{V}) \oplus \boldsymbol{\epsilon} \odot \boldsymbol{\sigma}(\mathbf{V})
	\end{aligned}
\end{equation*}} 
The $\mathbf{h}_{TM}$ is sampled from a posterior 
distribution $q(\mathbf{h}_{TM}|\mathbf{V})$ that is 
parameterized by mean $\boldsymbol{\mu}(\mathbf{V})$ and variance $\boldsymbol{\sigma}(\mathbf{V})$, 
generated by neural network. 
We call $\mathbf{h}_{TM}$ as a {\it document}-{\it topic}-{\it representation} ({DTR}), summarizing document semantics.  

(b) a softmax {\it decoder} $\hat {\bf V}$, i.e, $p(\mathbf{V}|\mathbf{h}_{TM})$ = $\prod_{i=1}^{N} p({\bf v_i} | \mathbf{h}_{TM})$ 
reconstructs the input document $\mathbf{V}$ by generating all words $\{{\bf v_i}\}$ independently:\vspace{-0.3cm}   

{\small \begin{equation*}\label{eq:nvdmdecoding}
	\begin{aligned}
		& p({\bf v_i}|\mathbf{h}_{TM}) = \frac{\mbox{exp}\{\mathbf{h}^T_{TM}\mathbf{U}_{:,i}+\mathbf{c_i}\}}{\sum_{j=1}^{Z}\mbox{exp}\{\mathbf{h}^T_{TM}\mathbf{U}_{:,i}+\mathbf{c_i}\}} \\
		&\mathcal{L}_{NVDM} = \mathbb{E}_{q(\mathbf{h}_{TM}|\mathbf{V})}\big[\log p(\mathbf{V}|\mathbf{h}_{TM})\big] -  \mbox{KLD}
	\end{aligned}
\end{equation*}}
where $\mathbf{U}$ $\in$ $\mathbb{R}^{K \times Z}$ and $\mathbf{c}$ $\in$ $\mathbb{R}^{Z}$ are decoding parameters, 
$\mathcal{L}_{NVDM}$ is the lower bound, 
i.e., $\log p(\mathbf{V})$ $\geq$ $\mathcal{L}_{NVDM}$ and 
$\mbox{KLD}$ = $\mbox{KL}[q(\mathbf{h}_{TM}|\mathbf{V}) || p(\mathbf{h}_{TM})]$ is the KL-Divergence  between 
the Gaussian posterior $q(\mathbf{h}_{TM}|\mathbf{V})$ and prior $p(\mathbf{h}_{TM})$ for $\mathbf{h}_{TM}$. 
During training, NVDM maximizes log-likelihood $\log p(\mathbf{V})$ $=$ $\sum_{{\bf h}_{TM}} p({\bf V}|{\bf h}_{TM}) p({\bf h}_{TM})$ 
by maximizing $\mathcal{L}_{NVDM}$ using stochastic gradient descent. 
See further details in \citet{DBLP:conf/icml/MiaoYB16}.

The \textbf{Neural Language Model} 
component (Figure~\ref{fig:topicBERTarchitecture}, right) is based on BERT \cite{BERT2019}. 
For a document $\mathcal{D}$ of length $N$, BERT first tokenizes the input sequence into a list of sub-word tokens ${\mathcal X}$ and then performs 
$O(N^{2}n_l)$ self-attention operations in $n_l$ encoding layers to compute its contextualized representation ${\bf o}_{CLS} \in \mathbb{R}^{{H}_B}$, 
extracted via a special token [CLS].  Here, $H_B$ is the number of hidden units. 
We use ${\bf o}_{CLS}$ to fine-tune BERT. 

\textbf{Complementary Learning}: \textit{TopicBERT} (Figure~\ref{fig:topicBERTarchitecture}) jointly 
performs neural topic and language modeling in a unified framework, where 
document-topic ${\bf h}_{TM}$ and contextualized ${\bf o}_{CLS}$  representations are first concatenated-projected 
to obtain a topic-aware contextualized representation $\mathbf{h}_p \in \mathbb{R}^{H_B}$ and then $\mathbf{h}_p$ is fed into a classifier:\vspace{-0.3cm}

{\small \begin{equation*}
	\begin{aligned}
	& \mathbf{h}_p = (\mathbf{h}_{TM}\;\circled{\mdoubleplus{}}\;\mathbf{o}_{CLS}) \cdot \mathbf{P}\\
	&p(y=y_l|\mathcal{D}) = 	\frac{\mbox{exp}\{\mathbf{h}^T_p\mathbf{Q}_{:,y}+\mathbf{b}_{y}\}}{\sum_{j=1}^{L}\mbox{exp}\{\mathbf{h}^T_p\mathbf{Q}_{:,y_j}+\mathbf{b}_{y_j}\}} \\
	& \mathcal{L}_{TopicBERT} = \alpha  \log p(y=y_l|\mathcal{D}) + (1-\alpha) \mathcal{L}_{NVDM}
	\end{aligned}
\end{equation*}}
where, $\mathbf{P} \in \mathbb{R}^{\hat{H} \times H_B}$ is the projection matrix, $\hat{H} = H + H_B$,
$\mathbf{Q} \in \mathbb{R}^{H_B \times L}$ \& $\mathbf{b} \in \mathbb{R}^{L}$ are classification parameters, 
$y_l \in \{y_1, ..., y_L\}$ is the true label for $\mathcal{D}$ and $L$ is the total number of labels. 
During training, the \textit{TopicBERT} maximizes the joint objective  $\mathcal{L}_{TopicBERT}$ 
with $\alpha \in (0, 1)$. 
Similarly, we extract ${\bf o}_{CLS}$ from DistilBERT \cite{DistilBERT2019} and the variant is named as {\it TopicDistilBERT}.

\subsection{TopicBERT: Efficient Fine-tuning}
Since the computation cost of BERT grows quadratically $O(N^2)$ with sequence length N and is limited to 512 tokens, 
therefore there is a need to deal with larger sequences. The {\it TopicBERT} model offers efficient fine-tuning by reducing the number 
of self-attention operations in the BERT component.  

In doing this, we split a document ${\mathcal D}$ into $p$ partitions each denoted by ${\mathcal D}'$ of length $N/p$.   
The NVDM component extracts document-topic representation ${\bf h}_{TM}$ efficiently for the input ${\mathcal D}$ and 
BERT extracts contextualized representation ${\bf o}_{CLS}$ for ${\mathcal D}'$, 
such that the 
self-attention operations are reduced by a factor of $p^2$ in each batch 
while still modeling all cross-partition dependencies within the complementary learning paradigm.    
Table \ref{table:seqlengthtopicBERT} illustrates the computation complexity of BERT vs TopicBERT and the efficiency achieved.  
\begin{table}[t]
	\centering
	\resizebox{0.49\textwidth}{!}{
		\begin{tabular}{c|c|c}
										& 		{\bf BERT} 					&		{\bf TopicBERT} 						\\ 						\hline
			Sequence length 				&			$N$						&		 $N/p$ 								\\ \cline{2-3}
			Time Complexity  			&		\multirow{2}{*}{$b(N^{2}H_B)n_l$}   			&			\multirow{2}{*}{$bKZ +b(N^{2}H_B/p^2)n_l $}		\\
			(batch-wise) 					&									&			\\ \cline{2-3}
			\#Batches					&		$n_b$						&			$p \times n_b$					\\ \cline{2-3}
			Time Complexity  			&		\multirow{2}{*}{$b(N^{2}H_B n_b)n_l $}		&		\multirow{2}{*}{$bKZn_b +b(N^{2}H_B n_b/p)n_l $} 	\\
			(epoch-wise) 				&									&			\\ \hline
		\end{tabular}
	}	
	\caption{Time complexity of BERT vs TopicBERT. 
Here, $b$: batch-size, $n_b$: \#batches and $n_l$: \#layers in BERT. 
Note, the compute cost of NVDM and self-attention operations as $KZ << (N^{2}H_B/p)n_l$.  In TopicBERT: $p = 1$ for complementary learning, and $p = \{2, 4, 8\}$ for complmenrary+efficient learning.} 
	\label{table:seqlengthtopicBERT}
\end{table}

\begin{table*}[t]
	\centering
	\setlength{\tabcolsep}{3pt}
	\renewcommand{\arraystretch}{1.17}
	\resizebox{0.895\textwidth}{!}{
		\begin{tabular}{cc|ccccc|ccccc}
			\hline
			& \multicolumn{1}{c|}{\multirow{2}{*}{\textbf{Models}}} & \multicolumn{5}{c|}{\textbf{Reuters8} (news domain)}  & \multicolumn{5}{c}{\textbf{Imdb} (sentiment domain)} \\
			& & \textit{F1} & \textit{Rtn} & $T_{epoch}$ & $T$ & $CO_{2}$  & \textit{F1} & \textit{Rtn} & $T_{epoch}$ & $T$ & $CO_{2}$  \\
			\hline
			\multirow{5}{*}{\rotatebox{90}{\small \textbf{baselines}}} 
			& \textit{CNN} & 0.852 $\pm$ 0.000 & 91.123\% & 0.007 & 0.340 & 14.51 &  0.884 $\pm$ 0.000 & 94.952\% & 0.201 & 2.010 & 85.83  \\
			& \textit{BERT-Avg} & 0.882 $\pm$ 0.000 & 94.331\% & - & 0.010 & 0.47  & 0.883 $\pm$ 0.000 & 94.844\% & - & 0.077 & 3.29  \\
			& \textit{BERT-Avg + DTR} & 0.867 $\pm$ 0.000 &  92.727\%  & - & 0.015 & 0.68  & 0.894 $\pm$ 0.000 & 96.026\% & - & 0.114 & 4.87  \\
			& \textit{DistilBERT} & 0.934 $\pm$ 0.003 & 99.893\% & 0.129 & 1.938 &  82.75 & 0.910 $\pm$ 0.003 & 97.744\% & {\bf 0.700} & {\bf 10.500} & {\bf 448.35}  \\
			& \textit{BERT} & 0.935 $\pm$ 0.012 & 100.00\% & 0.208 & 3.123 & 133.34  & 0.931 $\pm$ 0.002 & 100.00\% & 0.984 & 14.755 & 630.04  \\
			\hline
			\multirow{4}{*}{\rotatebox{90}{\small \textbf{proposal}}} 
			& \textit{TopicBERT-512} & \textbf{0.950} $\pm$ 0.005 & \textbf{101.60}\% & 0.212 & 3.183 & 135.93 & 0.934 $\pm$ 0.002 & 100.32\% & 1.017 & 15.251 & 651.22  \\
			& \textit{TopicBERT-256} & 0.942 $\pm$ 0.009 & 100.74\% & 0.125 & 1.870 & 79.85  & \textbf{0.936} $\pm$ 0.002 & \underline{\textbf{100.53}}\% & \underline{0.789} & \underline{11.838} & \underline{505.46} \\
			& \textit{TopicBERT-128} & 0.928 $\pm$ 0.015 & \underline{99.251}\% & \underline{\bf 0.107} & \underline{\bf 1.610} & \underline{\bf 68.76}  & 0.928 $\pm$ 0.002 & 99.677\% & 0.890 & 13.353 & 570.17  \\
			& \textit{TopicBERT-64} &  0.921 $\pm$ 0.006 & 98.502\% & 0.130 & 1.956 & 83.51  & 0.909 $\pm$ 0.015 & 97.636\% & 1.164 & 17.461 & 745.60 \\
			\hdashline
			& \textit{Gain (performance)} &  $\uparrow$ {\bf 1.604}\% & -& - & - & - &   $\uparrow$ {\bf 0.537}\% & -& - & - & - \\
			
			& \textit{Gain (efficiency)} &  - & {\bf 99.251}\% & $\downarrow${\bf 1.9}$\times$ & $\downarrow${\bf 1.9}$\times$ & $\downarrow${\bf 1.9}$\times$ &  - & {\bf 100.53}\% & $\downarrow${\bf 1.2}$\times$ & $\downarrow${\bf 1.2}$\times$ & $\downarrow${\bf 1.2}$\times$  \\\hline
			\hline
			& & \multicolumn{5}{c}{\textbf{20 Newsgroups (20NS)} (news domain)}  
			& \multicolumn{5}{c}{\textbf{Ohsumed} (medical domain)} \\
			& & \textit{F1} & \textit{Rtn} & $T_{epoch}$ & $T$ & $CO_{2}$  & \textit{F1} & \textit{Rtn} & $T_{epoch}$ & $T$ & $CO_{2}$  \\
			\hline
			\multirow{5}{*}{\rotatebox{90}{\small \textbf{baselines}}} 
			& \textit{CNN} & 0.786 $\pm$ 0.000 & 95.504\% & 0.109 & 1.751 & 74.76 &  0.684 $\pm$ 0.000 & 89.179\% & 0.177 & 7.090 & 302.74  \\
			& \textit{BERT-Avg}  & 0.692 $\pm$ 0.000 & 84.083\% & - & 0.037 & 1.58  & 0.453 $\pm$ 0.000 & 59.061\% & - & 0.094 & 4.01 \\
			& \textit{BERT-Avg + DTR}  & 0.731 $\pm$ 0.000 & 88.821\% & - & 0.051 & 2.18 & 0.543 $\pm$ 0.000 & 70.795\% & - & 0.191 & 8.16  \\
			& \textit{DistilBERT} & 0.816 $\pm$ 0.005 & 99.149\% & {\bf 0.313} & {\bf 4.700} & {\bf 200.69}  & 0.751 $\pm$ 0.006 & 97.913\% & {\bf 0.684} & {\bf 10.267} & {\bf 438.4}  \\
			& \textit{BERT} & 0.823 $\pm$ 0.007 & 100.00\% & 0.495 & 7.430 & 317.28  & 0.767 $\pm$ 0.002 & 100.00\% & 1.096 & 16.442 & 702.07 \\
			\hline
			\multirow{4}{*}{\rotatebox{90}{\small \textbf{proposal}}} 
			& \textit{TopicBERT-512}  & 0.826 $\pm$ 0.004 & 100.36\% & 0.507 & 7.606 & 324.76  & 
			\textbf{0.769} $\pm$ 0.005 & \textbf{100.26}\% & 1.069 & 16.036 & 684.75 \\
			& \textit{TopicBERT-256}  & 0.823 $\pm$ 0.016 & \underline{100.00}\%  & \underline{0.400} & \underline{5.993} & \underline{255.90}
			& 0.761 $\pm$ 0.001 & \underline{99.217}\% & \underline{0.902} & \underline{13.530} & \underline{577.73}  \\
			& \textit{TopicBERT-128}  & 0.826 $\pm$ 0.004 & 100.36\% & 0.444 & 6.666 & 284.64  &
			0.739 $\pm$ 0.006 & 96.349\% & 1.003 & 15.047 & 642.50 \\
			& \textit{TopicBERT-64} & \textbf{0.830} $\pm$ 0.002 & \textbf{100.85}\% & 0.605 & 9.079 & 387.66 & 0.711 $\pm$ 0.003 & 92.698\% & 1.334 & 20.008 & 854.34  \\
			\hdashline
			& \textit{Gain (performance)} &  $\uparrow$ {\bf 0.850}\% & - & - & - & - &  $\uparrow$ {\bf 0.260}\% & - & - & - & - \\
			& \textit{Gain (efficiency)} &  - & {\bf 100.00}\% & $\downarrow${\bf 1.2}$\times$ & $\downarrow${\bf 1.2}$\times$ & $\downarrow${\bf 1.2}$\times$  &  - & {\bf 99.217}\% & $\downarrow${\bf 1.2}$\times$ & $\downarrow${\bf 1.2}$\times$ & $\downarrow${\bf 1.2}$\times$  \\ \hline
	\end{tabular}}
	\caption{{\it TopicBERT} for document classification (macro-F1). 
\textit{Rtn}: Retention in \textit{F1} vs \textit{BERT}; 
$T_{epoch}$: average epoch time (in hours); 
$T$: $T_{epoch}$$\times$15 epochs;
$CO_2$: Carbon 
in $gram$ $eq$. (equation~\ref{eq:co2_estimation});
\textbf{bold}: Best (fine-tuned BERT-variant) in column;
\underline{underlined}: Most efficient {\it TopicBERT-x} vs {\it BERT};
Gain (performance): {\it TopicBERT-x} vs {\it BERT};
Gain (efficiency): \underline{underlined} vs {\it BERT}}  
	\label{table:baselinevsproposed1}
\end{table*}

\begin{figure*}[t]
	\centering
	\begin{subfigure}[b]{0.46\textwidth}
		\centering
		\includegraphics[scale=0.62,trim={0cm 0cm 0cm 0cm}, clip]{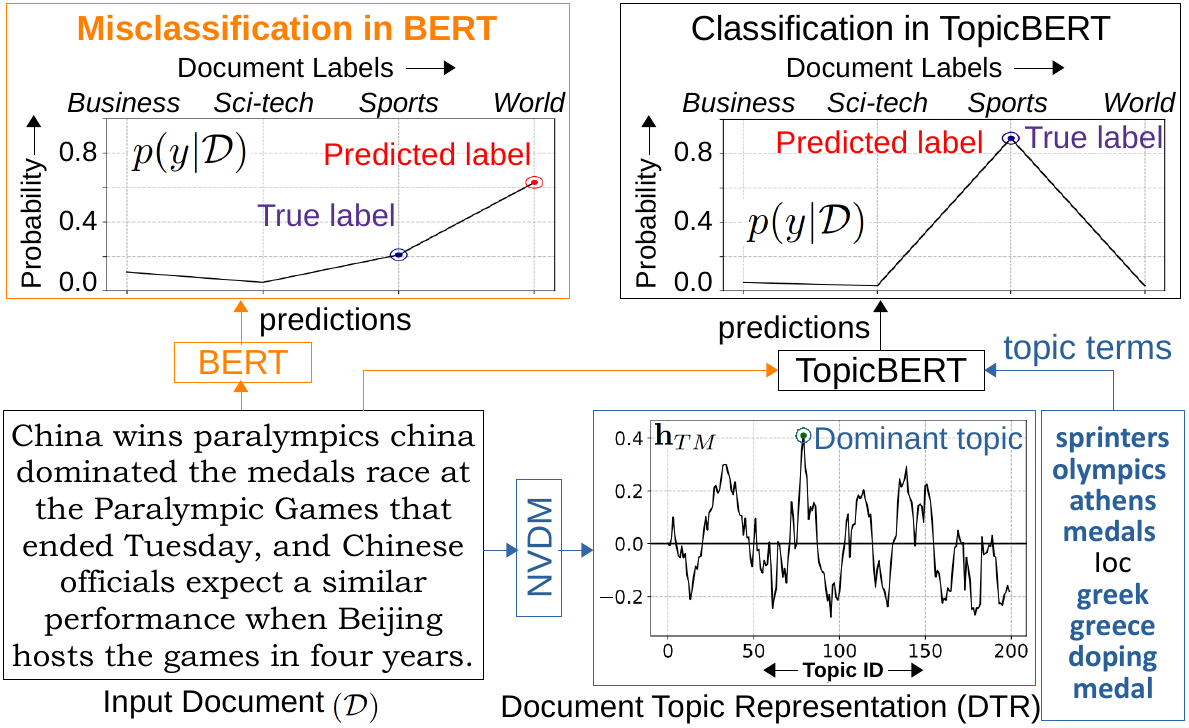}
	\end{subfigure}
	~~~~~~~
	\begin{subfigure}[b]{0.46\textwidth}
		\centering
		\includegraphics[scale=0.62,trim={0cm 0cm 0cm 0cm}, clip]{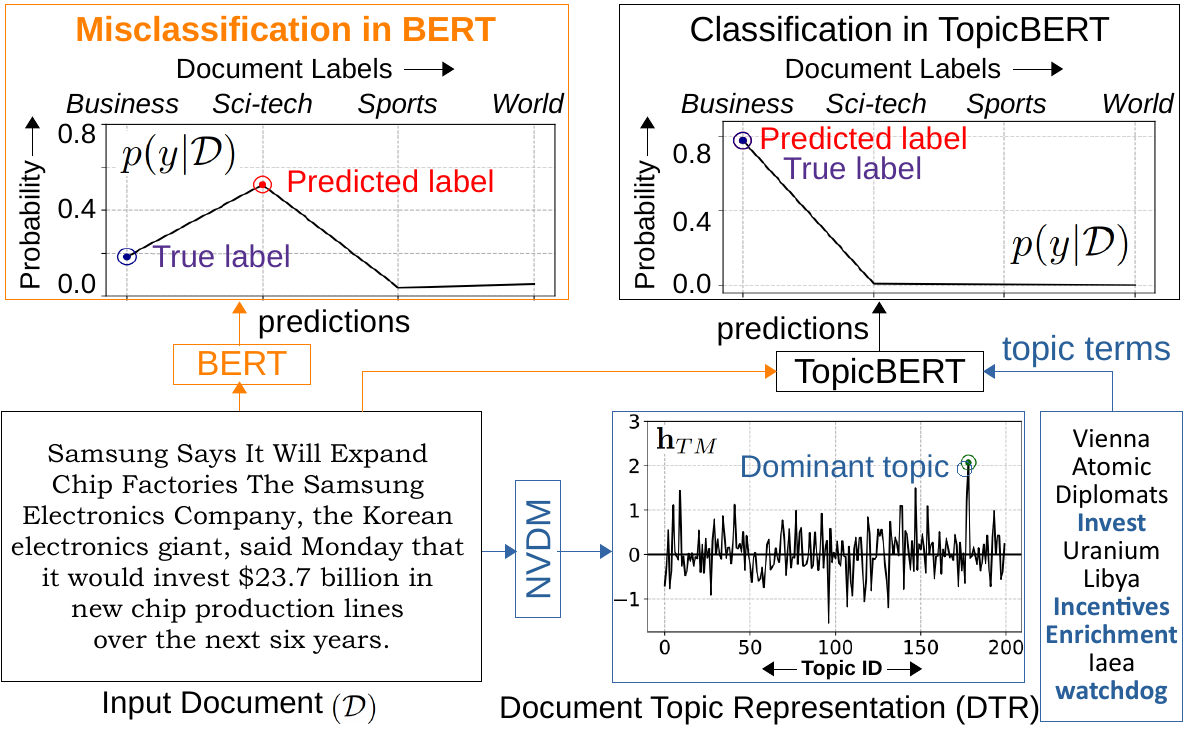}
	\end{subfigure}
	\caption{Interpretability analysis of document classification for AGnews dataset (for 2 different input documents): Illustration of document misclassification by \textit{BERT} and correct classification by \textit{TopicBERT} explained by the top key terms of dominant topic in DTR.}
	\label{fig:Interpretability_AGnews_sports_new}
\end{figure*}

\begin{figure*}[t]
	\centering
	\begin{subfigure}[b]{0.49\textwidth}
		\centering
		\includegraphics[scale=0.49,trim={0cm 0cm 0cm 0cm}, clip]{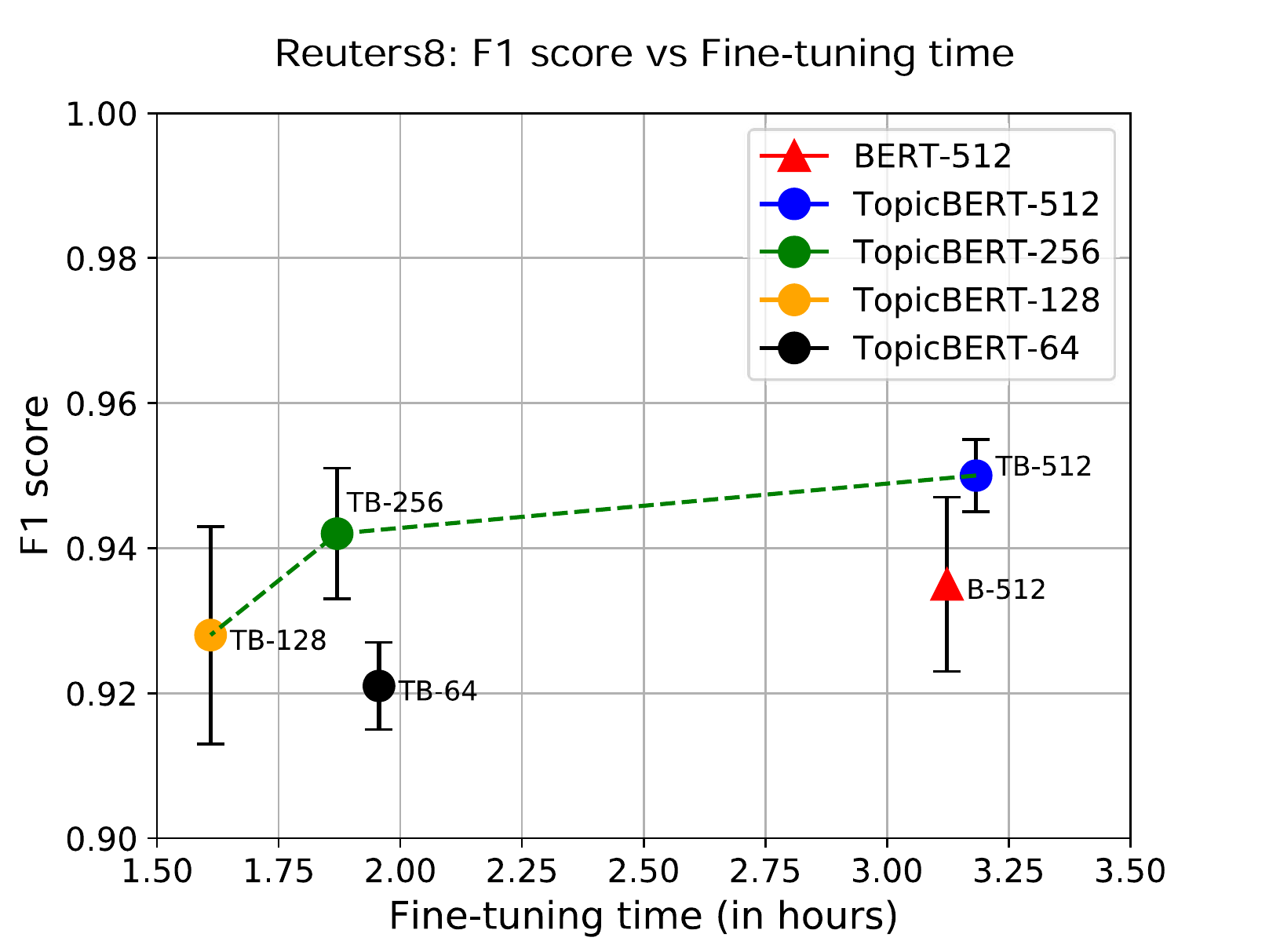}
	\end{subfigure}
	\begin{subfigure}[b]{0.49\textwidth}
		\centering
		\includegraphics[scale=0.49,trim={0cm 0cm 0cm 0cm}, clip]{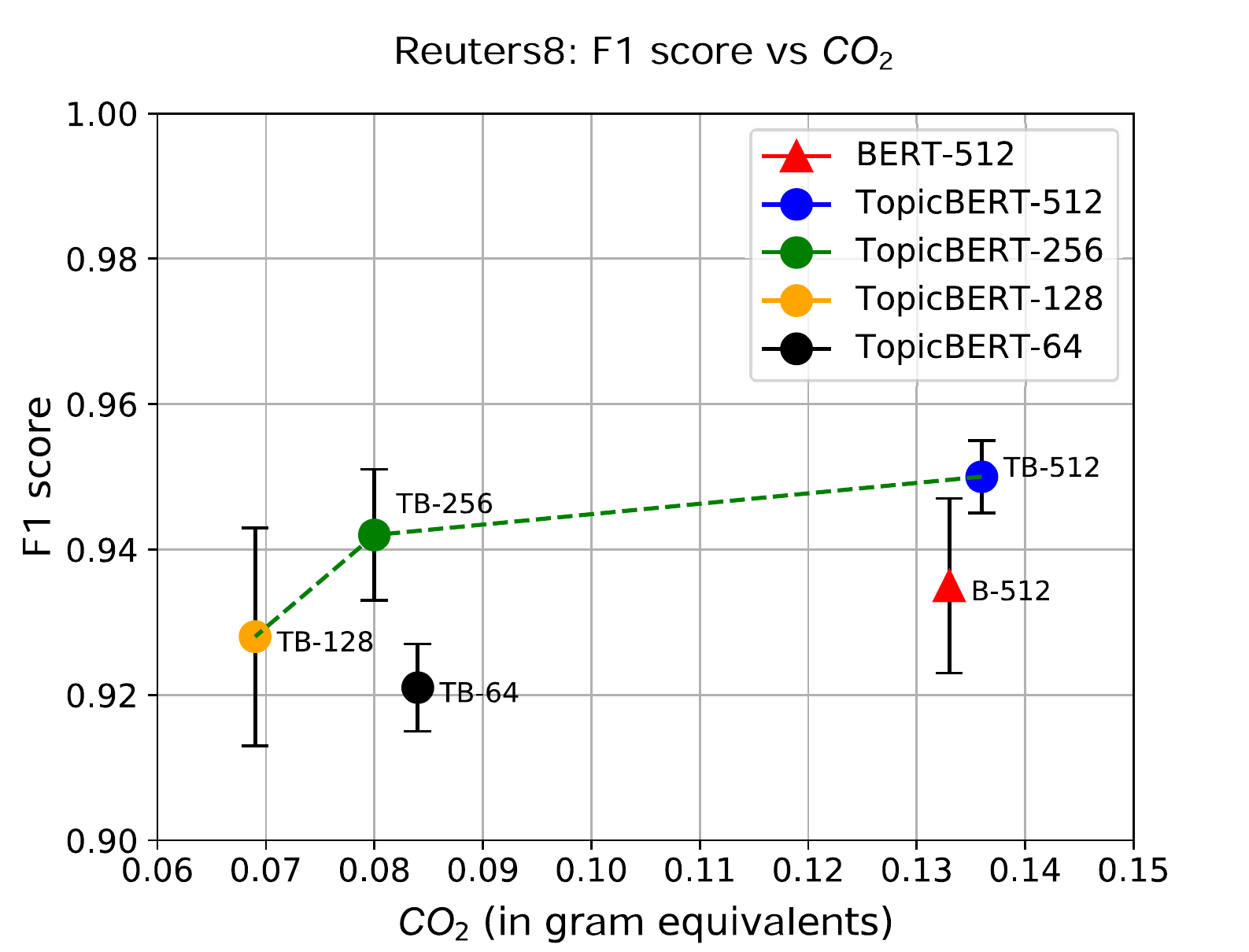}
	\end{subfigure}
	\caption{Pareto frontier analysis for Reuters8 dataset: \textit{F1} score vs Fine-tuning time (left) and F1 score vs $CO_2$ (carbon footprint) (right). Here green dashed line represents Pareto frontier connecting optimal solutions}
	\label{fig:reuters8_pareto}
\end{figure*}

\section{Experimental Results and Analysis}
\textbf{Datasets:} 
For document classification, we use 5 datasets ({\it Reuter8}, {\it Imdb}, {\it 20NS}, {\it Ohsumed}, {\it AGnews}) from several domains. (See \textit{supplementary} for data descriptions 
and experimental results of {\it AGnews}) 

\textbf{Baselines:} 
(a) \textit{CNN}~\cite{DBLP:conf/emnlp/Kim14}, 
(b) \textit{BERT-Avg}: Logistic classifier over the vector ${\bf \mathcal D}_{B}$ of a document obtained by averaging its contextualized word embeddings from {\it BERT}, 
(c) \textit{BERT-Avg}+{\it DTR}: 
Logistic classifier over concatenation(${\bf \mathcal D}_{B}$, {\it DTR}) where ${\it DTR} = {\bf h}_{TM}$ from pre-trained NVDM, i.e., no joint fine-tuning, 
(d) \textit{DistilBERT}~\cite{DBLP:journals/corr/abs-1910-01108}, (e) \textit{BERT} 
fine-tuned. 
We compare our extensions as: {\it TopicBERT} vs  {\it BERT} (below) and {\it TopicDistilBERT} vs {\it DistilBERT} (in {\it supplementary}).

\textbf{Experimental setup:} For {\it BERT} component, we split the input sequence $\mathcal{D}$ 
into $p$ equal partitions each of length $x = N_B/p$, where $N_B = 512$ (due to token limit of {\it BERT}) 
and $p$ $\in$ $\{1,2,4,8\}$ (a hyperparameter of {\it TopicBERT}). 
To avoid padding in the last partition, we take the last $x$ tokens of ${\mathcal D}$.
We run \textit{TopicBERT-x} (i.e., {\it BERT} component) for different sequence length ($x$) settings, 
where (a) $p =1 $, i.e., \textit{TopicBERT-512} denotes complementary fine-tuning, and 
(b) $p$ $\in$ $\{2,4,8\}$, i.e., \textit{TopicBERT}-\{$256$, $128$, $64$\} denotes complementary+efficient fine-tuning. 
Note, {\it NVDM} always considers the full-sequence. 
We execute 3 runs of each experiment on an NVIDIA Tesla T4 16 GB Processor to a maximum of 15 epochs.
Carbon footprint ($CO_2$) is computed as per equation~\ref{eq:co2_estimation}.
(See \textit{supplementary} for hyperparameters)

\textbf{Results}: Table \ref{table:baselinevsproposed1} illustrates gains in {\it performance} and {\it efficiency} of {\it TopicBERT}, respectively due to complementary and efficient fine-tuning. 
E.g. in Reuters8, {\it TopicBERT-512} achieves a gain of 1.6\% in $F1$ over {\it BERT} and also outperforms {\it DistilBERT}. 
In the efficient setup,  {\it TopicBERT-128} achieves a significant speedup of $1.9\times$ ($1.9\times$ reduction in $CO_2$) in fine-tuning while retaining ($Rtn$) 99.25\% of  \textit{F1} of {\it BERT}.
For IMDB and 20NS, {\it TopicBERT-256} reports similar performance to {\it BERT}, however with a speedup of $1.2\times$ and 
also outperforms {\it DistilBERT} in $F1$ though consuming similar time $T_{epoch}$.   
Additionally, {\it TopicBERT-512} exceeds {\it DistilBERT} in $F1$ for all the datasets. 
At $p$ = $8$, 
{\it TopicBERT-64} does not achieve expected efficiency perhaps due to saturated GPU-parallelization
(a trade-off in decreasing sequence length and increasing \#batches). 

Overall, {\it TopicBERT-x} achieves gains in: (a) {\it performance}: $1.604$\%, $0.850$\%, $0.537$\%, $0.260$\% and $0.319$\% in $F1$ for Reuters8, 20NS, IMDB, Ohsumed and AGnews (in \textit{supplementary}), respectively, and  
(b) {\it efficiency}: a speedup of 1.4$\times$ ($\sim$ $40$\%) and thus, a reduction of $\sim$ $40$\% in $CO_2$ 
over 5 datasets while retaining $99.9$\% of $F1$ compared to {\it BERT}. 
It suggests that the topical semantics improves document classification in {\it TopicBERT} (and {\it TopicDistilBERT}: a \underline{further} 1.55x speedup in {\it DistilBERT}) and its energy-efficient variants.   

{\bf Analysis (Interpretability)}:  
For two different input documents, 
Figure~\ref{fig:Interpretability_AGnews_sports_new} illustrates the misclassification by {\it BERT} and correct classification with explanation by {\it TopicBERT}, 
suggesting that the DTR (${\bf h}_{TM}$ of {\it NVDM}) improves document understanding. 
The {\it TopicBERT} extracts key terms of the dominant topic (out of 200) discovered by the {\it NVDM} component for each document. 
Observe that the topic terms explain the correct classification in each case. %
(See \textit{supplementary} for additional details and examples)

{\bf Analysis (Pareto Frontier)}: As shown in Table~\ref{table:baselinevsproposed1}, gains in \textit{TopicBERT} has been analyzed on two different fronts: (a) gain on the basis of \textit{performance} (\textit{F1} score), and (b) gain on the basis of \textit{efficiency} (Fine-tuning time/$CO_2$).
Figure~\ref{fig:reuters8_pareto} illustrates the following Pareto frontier analysis plots for Reuters8 dataset: (a) \textit{F1} score vs Fine-tuning time (left), and (b) \textit{F1} score vs $CO_2$ (right) to find the optimal solution that balances both fronts. 
Observe that the \textit{TopicBERT}-512 outperforms all other \textit{TopicBERT} variants and \textit{BERT} baseline (B-512) in terms of \textit{performance} i.e., \textit{F1} score. 
However, \textit{TopicBERT}-256 outperforms \textit{BERT}-512 in terms of both, \textit{performance} (\textit{F1} score) and \textit{efficiency} (Fine-tuning time/$CO_2$). 
Therefore, \textit{TopicBERT}-256 represents the optimal solution with optimal sequence length of 256 for Reuters8 dataset.

\section{Conclusion}
We have presented two novel architectures: {\it TopicBERT} and {\it TopicDistilBERT} for an improved and efficient (Fine-tuning time/$CO_2$) document classification, leveraging complementary learning of topic ({\it NVDM}) and language ({\it BERT}) models.   

\section*{Acknowledgments}
This research was supported by Bundeswirtschaftsministerium (bmwi.de), grant 01MD19003E (PLASS (plass.io)) at Siemens AG - CT Machine Intelligence, Munich Germany.

\bibliography{emnlp2020}
\bibliographystyle{acl_natbib}

\appendix

\section{Supplementary Material}

\subsection{$CO_2$: Carbon footprint estimation}
For Table 1, we estimate CO\textsubscript{2} for document classification tasks (BERT fine-tuning) considering 512 sequence length.  We  first estimate the frequency of BERT fine-tuning in terms of research activities and/or its application beyond. We estimate the following items:

1. Number of scientific papers based on BERT = 5532 (number of BERT citations to date: 01, June 2020)

2. Conference acceptance rate: 25\% (i.e., ~ 4 times the original number of submissions or research/application beyond the submissions)

3. Average number of datasets used = 5 

4. Average run-time of 15 epochs in fine-tuning BERT over 5000 documents (Reusters8-sized data) of maximum 512 sequence length = 12 hours on the hardware-type used

Therefore, using equation~\ref{eq:co2_estimation} in main paper,

CO\textsubscript{2}  estimate in fine-tuning BERT = 0.07 $\times$ (5532 $\times$ 4 $\times$  5)  $\times$ 12 $\times$ 0.61 kg eq. = 56,692 $\times$ 2,20462 lbs eq = 124,985 lbs eq.

\subsection{Data statistics and preprocessing}

Table~\ref{table:datastatistics} shows data statistics of 5 datasets used in complementary + finetuning evaluation of our proposed \textit{TopicBERT} model via Document Classification task. 20Newsgroups (20NS), Reuters8, AGnews are \textit{news} domain datasets, whereas Imdb and Ohsumed datasets belong to \textit{sentiment} and \textit{medical} domains respectively. For NVDM component, we preprocess each dataset and extract vocabulary $Z$ as follows: (a) tokenize documents into words, (b) lowercase all words, (d) remove stop words\footnote{we use NLTK tool to remove stopwords}, and (c) remove words with frequency less than $F_{min}$. Here, $F_{min} = 100$ for large datasets i.e., Imdb, 20NS, AGnews and Ohsumed, whereas $F_{min} = 10$ for Reuters8 which is a small dataset. 

\begin{table}[t]
	\centering
	\renewcommand{\arraystretch}{1.2}
	\resizebox{0.48\textwidth}{!}{
		\begin{tabular}{r|ccccccc}
			\hline
			\multirow{2}{*}{\textbf{Dataset}} & \textbf{Train} & \textbf{Dev}  & \textbf{Test} & \multirow{2}{*}{$Z$}  & \multirow{2}{*}{$L$} & \multirow{2}{*}{$N$} & \multirow{2}{*}{$b$} \\
			& \textbf{\#docs} & \textbf{\#docs}  & \textbf{\#docs} &  &  &  &  \\
			\hline
			Reuters8 &  4.9k  &  0.5k  &  2.1k  &  4813 &  8 & 512 & 4 \\
			Imdb &  20k &  5k &  25k &  6823 &  2 & 512 & 4 \\
			20NS &  9.9k &  1k &  7.4k &  4138 & 20 & 512 & 4 \\
			AGNews &  118k &  2k &  7.6k &  5001 &   4  & 128 & 32 \\
			Ohsumed$^\dagger$&  24k &  3k &  2.9k &  4553 & 20  & 512 & 4 \\
			\hline
	\end{tabular}}
	\caption{Preprocessed data statistics: \textbf{\#docs} $\rightarrow$ number of documents, k $\rightarrow$ thousand, $Z \rightarrow$ vocabulary size of NVDM, $L \rightarrow$ total number of unique labels, $N \rightarrow$ sequence length used for \textit{BERT} fine-tuning, $b \rightarrow$ batch-size used for BERT fine-tuning, ($^\dagger$) $\rightarrow$ multi-labeled dataset}
	\label{table:datastatistics}
\end{table}

\begin{table}[h]
	\centering
	\scriptsize
	\renewcommand{\arraystretch}{1.2}
	\resizebox{0.35\textwidth}{!}{
		\begin{tabular}{c|c}
			\hline
			Hyperparameter &  Value(s)\\
			\hline
			Learning rate  & \underline{0.001}, 0.05 \\ 
			Hidden size ($H$) & \underline{256}, 128 \\ 
			Batch size ($b$)  & 4, 32 \\ 
			Non-linearity ($g$) & sigmoid \\
			Sampling & \multirow{2}{*}{5, \underline{10}} \\
			frequency of $\mathbf{h}_{TM}$ & \\
			Number of & \multirow{2}{*}{50, \underline{100},{200}} \\
			topics ($K$) & \\
			\hline
	\end{tabular}}
	\caption{Hyperparameters search and optimal settings for NVDM component of \textit{TopicBERT} used in the experimental setup for document classification task.}
	\label{table:hypeparamnvdm}
\end{table}

\subsection{Experimental setup}

Table~\ref{table:hypeparamnvdm} and~\ref{table:hyperparambert} shows hyperparameter settings of NVDM and BERT components of our proposed \textit{TopicBERT} model for document classification task. We initialize BERT component with pretrained BERT-base model released by~\citet{BERT2019}. Fine-tuning of \textit{TopicBERT} is performed as follows: (1) perform pretraining of NVDM component, (2) initialize \textit{BERT} component with BERT-base model, (3) perform complementary + efficient fine-tuning, for 15 epochs, using joint loss objective:  

\begin{small}
	$\mathcal{L}_{TopicBERT} = \alpha \log p(y=y_l|\mathcal{D}) + (1-\alpha)\mathcal{L}_{NVDM}$
\end{small}

where, $\alpha \in \{0.1, 0.5, 0.9\}$. For CNN, we follow the experimental setup of Kim (2014). 

\begin{table*}[t]
	\centering
	\setlength{\tabcolsep}{3pt}
	\renewcommand{\arraystretch}{1.3}
	\resizebox{0.94\textwidth}{!}{
		\begin{tabular}{cc|ccccc|ccccc}
			\hline
			& \multicolumn{1}{c|}{\multirow{2}{*}{\textbf{Models}}} & \multicolumn{5}{c|}{\textbf{Reuters8} (news domain)}  & \multicolumn{5}{c}{\textbf{20NS} (news domain)} \\
			& & \textit{F1} & \textit{Rtn} & $T_{epoch}$ & $T$ & $CO_{2}$  & \textit{F1} & \textit{Rtn} & $T_{epoch}$ & $T$ & $CO_{2}$  \\
			\hline
			\multirow{2}{*}{\rotatebox{90}{\small \textbf{baselines}}} 
			& \textit{CNN} & 0.852 $\pm$ 0.000 & 91.123\% & 0.007 & 0.340 & 14.51 & 0.786 $\pm$ 0.000 & 95.504\% & 0.109 & 1.751 & 74.76  \\
			& \textit{DistilBERT} & 0.934 $\pm$ 0.003 & 100.00\% & 0.129 & 1.938 &  82.75 & 0.816 $\pm$ 0.005 & 100.000\% & 0.313 & 4.700 & 200.69  \\
			\hline
			\multirow{3}{*}{\rotatebox{90}{\small \textbf{proposal}}} 
			& \textit{TopicDistilBERT-512} & 0.941 $\pm$ 0.007 & 100.75\% & 0.132 & 1.976 & 84.37 & {\bf 0.820} $\pm$ 0.000 & {\bf 100.49}\% & 0.320 & 4.810 & 205.38  \\
			& \textit{TopicDistilBERT-256} & {\bf 0.943} $\pm$ 0.006 & \underline{\bf 100.96}\% & \underline{\bf 0.085 }& \underline{\bf 1.272} & \underline{\bf 54.31}  & 0.802 $\pm$ 0.000 & \underline{98.284}\% & \underline{\bf 0.190} & \underline{\bf 2.850} & \underline{\bf 121.69} \\
			& \textit{TopicDistilBERT-128} & 0.911 $\pm$ 0.012 & 97.573\% & 0.096 & 1.444 & 61.66  & 0.797 $\pm$ 0.000 & 97.671\% & 0.387 & 5.800 & 247.66  \\
			\hdashline
			& \textit{Gain (performance)} &  $\uparrow$ {\bf 0.964}\% & -& - & - & - &   $\uparrow$ {\bf 0.490}\% & -& - & - & - \\
			
			& \textit{Gain (efficiency)} &  - & {\bf 100.96}\% & $\downarrow${\bf 1.5}$\times$ & $\downarrow${\bf 1.5}$\times$ & $\downarrow${\bf 1.5}$\times$ &  - & {\bf 98.284}\% & $\downarrow${\bf 1.6}$\times$ & $\downarrow${\bf 1.6}$\times$ & $\downarrow${\bf 1.6}$\times$  \\\hline
	\end{tabular}}
	\caption{{\it TopicDistilBERT} vs {\it DistilBERT} for document classification (macro-F1) in complementary ({\it TopicDistilBERT}-512) and efficient ({\it TopicDistilBERT}-\{256, 128\}) learning setup. Here, 
		\textit{Rtn}: Retention in \textit{F1} vs \textit{BERT}; 
		$T_{epoch}$: average epoch time (in hours); 
		$T$: $T_{epoch}$$\times$15 epochs;
		$CO_2$: Carbon footprint 
		in $gram$ $eq$. (equation~\ref{eq:co2_estimation});
		\textbf{bold}: Best (fine-tuned DistilBERT-variant) in column;
		\underline{underlined}: Most efficient {\it TopicDistilBERT-x} vs {\it DistilBERT};
		Gain (performance): {\it TopicDistilBERT-x} vs {\it DistilBERT};
		Gain (efficiency): \underline{underlined} vs {\it DistilBERT}}  
	\label{table:baselinevsproposedsuppl}
\end{table*}

\begin{table}[h]
	\centering
	\scriptsize
	\renewcommand{\arraystretch}{1.2}
	\resizebox{0.40\textwidth}{!}{
		\begin{tabular}{c|c}
			\hline
			Hyperparameter &  Value(s)\\
			\hline
			Learning rate*   & 2e-5\\ 
			Hidden size ($H_B$) & 768 \\ 
			Batch size ($b$)  & [4, 32] \\ 
			Non-linearity* & gelu \\
			Maximum sequence & [512, 256, \\
			length ($N$) & 128, 64, 32$^\ddagger$] \\
			Number of & \multirow{2}{*}{12} \\
			attention heads* & \\
			Number of &  \multirow{2}{*}{12} \\
			encoder layers* ($n_l$) & \\
			Vocabulary size* & 28996 \\
			Dropout probability* & 0.1 \\
			$\alpha$ & [0.1, 0.5, \underline{0.9}] \\
			\hline
		\end{tabular}
	}
	\caption{Hyperparameters search and optimal settings for BERT component of \textit{TopicBERT} used in the experimental setup for document classification. $^\dagger \rightarrow$ additional hyperparameter introduced for joint modeling in \textit{TopicBERT}, $^\ddagger \rightarrow$ $N = 32$ is only used for AGnews dataset, (*) $\rightarrow$ hyperparameter values taken from pretrained BERT-base model released by~\citet{BERT2019}.}
	\label{table:hyperparambert}
\end{table}

\begin{table}[t]
	\centering
	\setlength{\tabcolsep}{3pt}
	\renewcommand{\arraystretch}{1.25}
	\resizebox{0.48\textwidth}{!}{
		\begin{tabular}{cc|ccccc}
			\hline
			& \multicolumn{1}{c|}{\multirow{2}{*}{\textbf{Models}}} & \multicolumn{5}{c}{\textbf{AGnews}} \\
			& & F1 & Rtn & $T_{epoch}$ & T & $CO_{2}$  \\
			\hline
			\multirow{5}{*}{\rotatebox{90}{\small \textbf{baselines}}} 
			& \textit{CNN} & 0.916 $\pm$ 0.000 & 97.447\% & 0.131 & 0.921& 393.25 \\
			& \textit{BERT-Avg} & 0.903 $\pm$ 0.000 & 96.064\% & - & 0.075 & 3.20 \\
			& \textit{BERT-Avg + DTR} & 0.913 $\pm$ 0.000 & 97.128\% & - & 0.105 & 4.48 \\
			& \textit{DistilBERT-x} & 0.941 $\pm$ 0.001 & 100.10\% & \textbf{0.491} & \textbf{7.361} & \textbf{314.31} \\
			& \textit{BERT-x} & 0.940 $\pm$ 0.001 & 100.00\% & 0.952 & 14.281 & 609.80 \\
			\hline
			\multirow{3}{*}{\rotatebox{90}{\small \textbf{proposal}}} 
			& \textit{TopicBERT-128} & 0.942 $\pm$ 0.003 & 100.21\% & 1.004 & 15.065 & 643.27  \\
			& \textit{TopicBERT-64} & \textbf{0.943} $\pm$ 0.002 & \underline{\textbf{100.31}\%} & \underline{0.723} & \underline{10.838} & \underline{462.78}  \\
			& \textit{TopicBERT-32} & 0.938 $\pm$ 0.001 & 99.78\% & 0.846 & 12.688 & 541.66  \\
			\hdashline
			& \textit{Gain (performance)} &  $\uparrow$ {\bf 0.319} \% & - & - & - & - \\
			& \textit{Gain (efficiency)} &  - & {\bf 100.31}\% & $\downarrow$ {\bf 1.3} $\times$ & $\downarrow$ {\bf 1.3} $\times$ & $\downarrow$ {\bf 1.3} $\times$ \\
			\hline
	\end{tabular}}
	\caption{\textit{TopicBERT} for document classification (macro-F1) for AGnews dataset. 
		\textit{Rtn}: Retention in \textit{F1} vs \textit{BERT}; 
		$T_{epoch}$: average epoch time (in hours); 
		$T$: $T_{epoch}$$\times$15 epochs;
		$CO_2$: Carbon footprint 
		in $gram$ $eq$. (equation~\ref{eq:co2_estimation});
		\textbf{bold}: Best (fine-tuned BERT-variant) in column;
		\underline{underlined}: Most efficient {\it TopicBERT-x} vs {\it BERT};
		Gain (performance): {\it TopicBERT-x} vs {\it BERT};
		Gain (efficiency): \underline{underlined} vs {\it BERT}}  
	\label{table:resultsagnews}
\end{table}

\subsection{Results of TopicBERT for AGnews}

Table~\ref{table:resultsagnews} shows gains in \textit{performance} and \textit{efficiency} of \textit{TopicBERT} vs \textit{BERT} for AGnews dataset. \textit{TopicBERT} achieves: (a) a gain of 0.3\% in \textit{F1} (\textit{performance}) compared to \textit{BERT}, and (b) a significant speedup of 1.3$\times$ over \textit{BERT} while retaining (\textit{Rtn}) 100\% of \textit{F1} (\textit{performance}) of \textit{BERT} at the same time. This gain arises due to the improved document understanding using complementary topical semantics, via NVDM, in \textit{TopicBERT} and its energy efficient versions.

\subsection{TopicDistilBERT vs DistilBERT}

Table \ref{table:baselinevsproposedsuppl} reports scores of {\it TopicDistilBERT} vs {\it DistilBERT} for two datasets (Reuters8 and 20NS).  
We follow the similar schemes of sequence lengths (512, 256 and 128) to evaluate the performance of the 
(a) complementary learning via {\it TopicDistilBERT-512} vs {\it DistilBERT}, and
(b) efficient learning via {\it TopicDistilBERT-\{256, 128\}} vs {\it DistilBERT}. 

For Reuters8 in complementary setup, {\it TopicDistilBERT-512} achieves a gain (0.941 vs 0.934)  in $F1$ over {\it DistilBERT}.  
In the efficient setup,  {\it TopicDistilBERT-256} achieves a significant speedup of $1.5\times$ ($1.5\times$, i.e., $\sim$50\% reduction in CO\textsubscript{2}) 
in fine-tuning while retaining ($Rtn$) 100.96\% of \textit{F1} of {\it DistilBERT}.

For 20NS in complementary setup, {\it TopicDistilBERT-512} achieves a gain (0.820 vs 0.816)  in $F1$ over {\it DistilBERT}.  
In the efficient setup, {\it TopicDistilBERT-256} achieves a speedup of $1.6\times$ ($1.6\times$, i.e., $\sim$60\% reduction in CO\textsubscript{2}).
   
Additionally, {\it TopicBERT-512} exceeds {\it DistilBERT} in $F1$ for the two datasets. 
At $p$ = $4$, 
{\it TopicDistilBERT-128} does not achieve expected efficiency perhaps due to saturated GPU-parallelization
(a trade-off in decreasing sequence length and increasing \#batches) 
and therefore, we do not partition further. 

Overall, {\it TopicDistilBERT-x} achieves gains in: (a) {\it performance}: $0.964$\%, and $0.490$\% in $F1$ for Reuters8 and 20NS, respectively, and  
(b) {\it efficiency}: a speedup of 1.55$\times$ ($\sim$ $55$\%) and thus, a reduction of $\sim$ $55$\% in CO\textsubscript{2} 
over 2 datasets while retaining $99.6$\% of $F1$ compared to {\it DistilBERT} baseline model. 

It suggests that the topical semantics improves document classification in {\it TopicDistilBERT} (and {\it TopicBERT}) and its energy-efficient variants.   
Based on our two extensions: {\it TopicBERT} and {\it TopicDistilBERT}, we assert that our proposed approaches of complementary learning (fine-tuning) are {\it model agnostic} of BERT models. 

\begin{figure}[t]
	\centering
	\includegraphics[scale=0.62,trim={0cm 0cm 0cm 0cm}, clip]{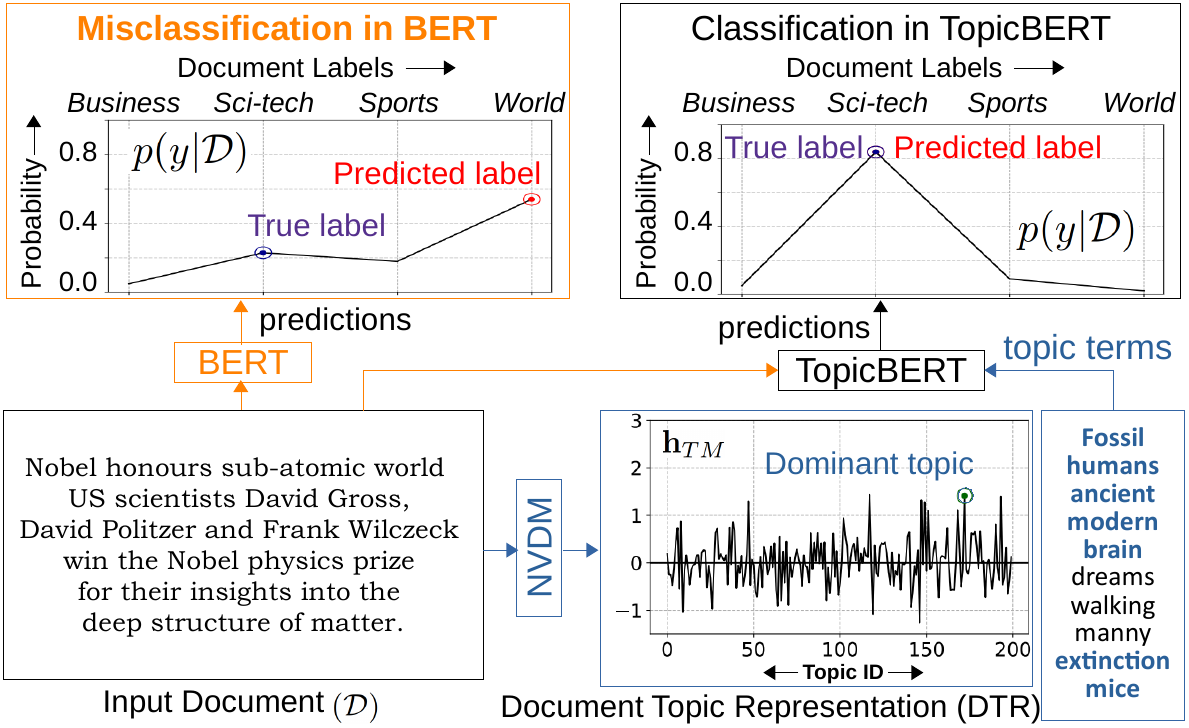}
	\caption{Interpretability analysis of document classification for AGnews dataset (for 2 input documents): Illustration of document misclassification by \textit{BERT} model and correct classification by \textit{TopicBERT} explained by the top key terms of dominant topic in DTR.}
	\label{fig:agnews_interpretability}
\end{figure}

\subsection{Interpretability Analysis in TopicBERT}
To analyze the gain in \textit{performance} (\textit{F1} score) of \textit{TopicBERT} vs \textit{BERT}, 
Figure~\ref{fig:agnews_interpretability} shows document label misclassifications due to \textit{BERT} model. However, \textit{TopicBERT} model is able to correctly predict the labels using document topic representation (DTR) which explains the correct predictions by the top key terms of the dominant topic discovered.

\end{document}